\documentclass[sigconf]{acmart}
\usepackage{graphicx} 
\usepackage{url}
\usepackage{float}
\usepackage{hyperref}
\usepackage{subfigure}
\usepackage{epsfig}
\usepackage[ruled,vlined,linesnumbered]{algorithm2e}
\newcommand{\eat}[1]{}

\begin{document}

\title{A Declarative Query Language for Scientific Machine Learning}

\author{Hasan M. Jamil}
\affiliation{%
  \country{University of Idaho, USA}
}
\email{jamil@uidaho.edu}

\renewcommand{\shortauthors}{Hasan Jamil}
\renewcommand{\shorttitle}{MQL}

\begin{abstract}
The popularity of data science as a discipline and its importance in the emerging economy and industrial progress dictate that machine learning be democratized for the masses. This also means that the current practice of workforce training using machine learning tools, which requires low-level statistical and algorithmic details, is a barrier that needs to be addressed. Similar to data management languages such as SQL, machine learning needs to be practiced at a conceptual level to help make it a staple tool for general users. In particular, the technical sophistication demanded by existing machine learning frameworks is prohibitive for many scientists who are not computationally savvy or well versed in machine learning techniques. The learning curve to use the needed machine learning tools is also too high for them to take advantage of these powerful platforms to rapidly advance science. In this paper, we introduce a new declarative machine learning query language, called {\em MQL}, for naive users. We discuss its merit and possible ways of implementing it over a traditional relational database system. We discuss two materials science experiments implemented using MQL on a materials science workflow system called MatFlow.
\end{abstract}



\date{October 2023}

\def\BibTeX{{\rm B\kern-.05em{\sc i\kern-.025em b}\kern-.08emT\kern-.1667em\lower.7ex\hbox{E}\kern-.125emX}}

\copyrightyear{2024}
\acmYear{2024}
\setcopyright{acmcopyright}
\acmPrice{15.00}
\acmDOI{10.1145/xxxxxxxxxx}
\acmISBN{xxxxxxxxxxxx}

\acmArticle{4}
\acmPrice{15.00}


\maketitle

\section{Introduction}
\label{intro}

In this paper, we ask the question, how difficult it is to design a declarative query language for machine learning (ML) analysis by pointing to how difficult and arcane it is to write code segments in popular ML platforms such as SciKit-Learn, Pytorch, R or TensorFlow by non ML experts? By declarative, we mean that if a language for ML that is as simple and as powerful as SQL can be designed, can it perform the most complex analysis a modern ML algorithm can?

The current state of ML is not accessible to most of potential users of data science \cite{HershAHG22s}, scientists in particular, and we concur with many researchers who believe that a significant barrier exists towards exploiting ML without a declarative platform \cite{MolinoR21}. In the absence of a language similar to SQL, it is extremely difficult and unlikely for naive users and scientists alike to  comprehend, let alone devise, a simple regression analysis code fragment easily executable on a machine. For example, the process to perform a clustering analysis \cite{Ranta2018} (or classification \cite{McCaffrey2021}) on the Boston housing dataset on Kaggle \cite{Perera2018} is by no means an easy task even for a good computational scientist, without adequate proficiency in regression analysis. It requires exploratory data analysis, principal component analysis, and more to get a sense of the data and to make a decision about the number of clusters that are appropriate, most of which can also be performed by a smart algorithm. Then there is the issue of accuracy and selection of the best model for the analysis \cite{Soper22,AbouloifaB22,Hosseini2023}. 

The natural question then is, are all these details necessary, at least most of the time? Could these analysis algorithms be chosen by a query processor from an abstract request for prediction, clustering or classification based on the properties of the data sets the same way relational database engines select join algorithms, aggregate function algorithms or procedures for OLAP functions? Could optimization be possible and decided by query processors in ways analogous to SQL engines?

While we do not currently have all the answers, we believe that the starting point should be the development of a suitable declarative query language for ML that will be simple in spirit and expressive enough to be able to support most, if not all, ML analysis needs on tabular data. To that end, our goal in this paper is to introduce an ML query language, called {\em MQL} (stands for Machine learning Query Language), capable of supporting three basic classes of ML tasks - prediction, classification and clustering. We stagger the language constructs in three tiers -- data preparation (or wrangling), model construction, and ML analysis. These language constructs have distinct semantics and no inherent inter-dependencies.

Finally, as data science becomes mainstream \cite{AodhaSBTGJ14s} and scientists increasingly rely on ML tools almost in every areas of science and engineering (e.g., \cite{AkhtarGKHAZKA23,VermaKG22,QiBRT22,MadiyarovTAMY23}), efforts are needed to lower the entry barriers to ML tools for scientists. As various user aids emerge (e.g., visual interfaces \cite{LuongMBR20,RojoRR018}, natural language interfaces \cite{RatnerBEFWR20}, ML tools \cite{BeltranBCFMS24,ArostegiM0S21s,PietroD19}, arresting ML application development costs \cite{Coop2021,arxiv.2209.07124,lorica2018state} are becoming imperative. New ways must be found to reduce access and application development costs involving AI and ML. Declarative query languages reduce the access barrier, and thus costs of data analysis by allowing minimally trained experts to use ML. We believe, declarative languages also are more amenable to automated code synthesis, e.g., natural language interfaces to scientific applications that can be constructed fully autonomously from software specifications, again expressed in natural language.

In the sections to follow, we first present MQL's syntax and semantics, and then discuss its merit over the contemporary declarative languages. Using two materials science experiments in our lab, we highlight how easy cost-inspiring it is to use MQL for scientific application using tabular data and traditional ML. While more research is needed to make MQL more expressive and powerful to support more sophisticated ML frameworks such as deep learning, the current edition of MQL paves the way for further extensions.

\section{Related Work}
\label{related}

The main purpose of a declarative language to reduce the human-machine interfacing barriers by making machine instructions simple and easy to conceptualize. An all time great example of declarative languages is SQL. While this definition of declarativity is subjected to interpretation, the essence should remain. From this standpoint, a simple language for ML has to be highly abstract, and should support the so called naive users' use of ML tasks having only conceptual and rudimentary knowledge of this technology while the machines assume the bulk of the technical underpinnings and efficiency concerns \cite{Vaithyanathan09}. Given that ML tasks are complex, involved, and require subject expertise, meeting such levels of abstraction requirements in a human-machine interfacing language, or query language, is undoubtedly a tall order.

Nonetheless, several attempts were made to simplify the use of ML technologies for the masses. Among them, AutoML \cite{KarmakerHSXZV22} maybe the most prominent effort of all. While challenges remain \cite{DeyGD23s}, the emergence of large language models appear to address many of these challenges to some extent \cite{abs-2306-08107} toward democratizing ML. AutoML, or Automated Machine Learning, is a set of techniques aimed at automating the process of building ML models. The basic idea behind AutoML is to make ML more accessible to users with limited ML expertise by automating some of the complex and time-consuming tasks, such as data preprocessing, feature engineering,  model selection, and hyperparameter optimization, involved in model development. By automating these tasks, AutoML aims to reduce the amount of manual effort required to build and deploy ML models, making it easier for non-experts to leverage the power of ML in their applications.

Variants of the ideas behind AutoML are also being investigated. Among them, MLBase \cite{KraskaTDGFJ13s} attempted to help automating the pipeline by proposing a declarative language and an optimizer to lend a hand in balancing the efficiency aspects of declarativity that usually delegates this responsibility to the system. Despite the design goal, the language they support appears to retail procedural features still and is not abstract enough compared to languages such as SQL to have a wider appeal.

An early effort to develop a simpler ML front-end was a natural language interpreter called WOLFE \cite{SinghRHNR15}. In approaches such as WOLFE, query understanding and mapping the intent into some form of executable code is employed, in WOLFE the code is written in TensorFlow. A similar translational approach is used in languages such as sql4ml \cite{abs-1907-12415}, ML2SQL/MLearn \cite{SchuleBVKG019,SchuleBKG019a}, P6 \cite{LiM21}, MLog \cite{LiCCWZ17}, Datalog \cite{WangWLGDZ21}, and Dyna \cite{VieiraFFKE17}. The popularity of translational implementation of declarative languages to ML frameworks such as TensorFlow, PyTorch or SciKit Learn is not by accident. Rather it is convenience and a desire to leverage the community investments in powerful and a large body of algorithms for ML over a few decades. While a more powerful end-to-end ML systems are developed and mature, such as SystemML \cite{BoehmDEEMPRRSST16}, SystemDS \cite{BoehmADGIKLPR20s}, EndToEndML \cite{abs-2403-18203}, Merlion \cite{BhatnagarKLLYCSASWSJGSKMCZZXSHW23}, VeML \cite{abs-2304-13037}, and Relax \cite{abs-2311-02103}, we believe these translation grounded systems will continue to play a major role in democratizing ML.

In sql4ml, an SQL like ML instruction is mapped to TensorFlow script. However, the CREATE VIEW abstraction conceived can do little to hide the subject expertise users need to state the analysis needs defeating the purpose of a declarative language for ML. The MLearn system \cite{SchuleBKG019a} also do so using its ML2SQL mapping approach \cite{SchuleBVKG019}. The operator creation based approach is tedious and rests significant domain knowledge burden on the users. Dyna and P6 systems bring optimization and visualization capabilities into the declarative ML landscape. A more systematic investigation to deal with performance of AutoML engines show that ML pipelines efficiency can be improved using a predictive model \cite{ZhangLSOBF23}. And in cases where automation is difficult, a human-in-the-loop approach may also help \cite{XanthopoulosTCS20}.

In MQL, however, we adopt a SQL-Centric approach, as opposed to DSL-Centric or UDF-Centric approaches \cite{0001EPR16} and propose an entirely new declarative language for ML in the spirit of SQL even though we too rely on a translational approach to its implementation. While we present a mapping to SciKit Learn for its implementation as a proof of concept, we note that more needs to be done to make MQL a viable system for serious ML platform. In sections \ref{results} and \ref{discussion}, we will present MQL's capabilities in scientific computing, and possible improvements respectively to elaborate further. 

\section{Machine Learning Query Language}

With the intent to stay close to an SQL-like language, called MQL, we propose a syntax similar to SQL and design lower level operational procedures to assign a semantics to the declarative statements of MQL. MQL retains part of SQL flavor to leverage the community knowledge of SQL and reduce cognitive overload. 

\subsection{Syntax of MQL}

Similar to SQL, MQL supports two basic statements -- the {\sf GENERATE} statement for querying tables and {\sf CONSTRUCT} statement for creating a ML model. While the {\sf GENERATE} statement is able to exploit an existing model, it can also operate without one by generating its own model.

\subsubsection{The GENERATE Statement}

GENERATE statement stands at the level of SQL's SELECT statement and is the main workhorse of MQL. It operates on tabular data to make predictions, categorize objects and group sets of objects into bins. It has five basic clauses -- an ML class selection (one of PREDICTION, CLASSIFICATION and CLUSTER), optional object labeling, feature selection, a data set, a filter condition over the data set, and an input table of unknown cases (test set).

\begin{verbatim}
   GENERATE [DISPLAY OF]
   PREDICTION v [OVER s] |
   CLASSIFICATION INTO L1, L2, ..., Lp [OVER s] |
   CLUSTER OF k
   [USING MODEL ModelName | ALGORITHM AlgorithmName]
   [WITH MODEL ACCURACY P]
   [LABEL B1, B2, ..., Bm]
   [FEATURES A1, A2, ..., An
   FROM r1, r2, ..., rq
   WHERE c]
\end{verbatim}

In the above statement, $r_l$ is a table over the scheme $R_l$, $c$ is a Boolean condition, $A_i \in \cup_l R_l$, $s$ is a table over the scheme $\cup_j B_j \bigcup \cup_i A_i$, k is an integer, and $v \in  \cup_l R_l$, $L_k \in  dom(X)$\footnote{$dom(X)$ is the set of elements in the domain of the column $X$, and $X \in  \cup_i A_i$.}. In this statement and in all the MQL statements, the vertical bar ($|$) means exclusive OR, and the square bracket ($[]$) means optional.

$v$ in the PREDICTION clause is the target variable, and $A_i$s are the features. The optional LABEL clause identifies attributes $B_j$ as the object identifiers for all the three ML tasks. The CLASSIFICATION clause classifies each object $\cup_j B_j$ into one of $L_k$ categories. The $k$ in CLUSTER clause is an integer expression that can include SQL aggregate functions over the tables $r_l$. Finally, the optional USING clause is meant to either use an existing model (MODEL option) generated using the CONSTRUCT clause (discussed next), or a specific ML algorithm (ALGORITHM option) for the generation of the model. As in SQL, WHERE is an optional clause, but unlike SQL, FROM is required. The OVER clause supplies the unknown test dataset over the scheme $A_i \cup B_j$. The ACCURACY option accepts a threshold within the interval (0,1).

\subsubsection{CONSTRUCT Statement}

To create an explicit model, MQL uses the CONSTRUCT statement below. It stands at a level similar to SQL's CREATE TABLE statement, but is at the data level and more functional. It is able to generate a default model for prediction, classification or clustering, optionally using a specific algorithm for supervised or unsupervised learning. The TRAIN ON parameter N (similarly TEST ON) is an integer value less than the cardinality of the table $r$, and can be expressed as an expression, possibly using SQL aggregate functions. While the expression for $M$ should be such that $N+M \leq |r|$ (where $|r|=|r_1| \times |r_2| \times \ldots \times |r_n|$), MQL will not object if the condition $N+M \leq |r|$ is not met and will assign the eventual semantics entailed by these two expressions. In this statement, $A_i$ is the feature vector over which the model is created.

\begin{verbatim}
   CONSTRUCT ModelName [AS SUPERVISED | UNSUPERVISED]
   FOR PREDICTION v |
   CLASSIFICATION INTO L1, L2, ..., Lp |
   CLUSTER OF k
   [USING AlgorithmName]
   [WITH MODEL ACCURACY P]
   TRAIN ON N TEST ON M
   FEATURES A1, A2, ..., An
   FROM r1, r2, ..., rn
   WHERE c
\end{verbatim}

\subsubsection{The INSPECT Statement}

The INSPECT statement is similar to the UPDATE statement of SQL and helps editing or wrangling the tables. For attributes $A_i$, it allows the values in these columns to be categorized, missing values predicted, convert categories to continuous values and eliminate duplicate rows. This statement affords MQL the power to manipulate a table to make it suitable for a potential learning task fully autonomously by a smart pre-processing engine. INSPECT returns a table with a scheme of a relation reflective of the resulting table in the FROM clause.

\begin{verbatim}
   INSPECT A1 [CATEGORIZE INTO L1, L2, ..., Lx |
   IMPUTE | NUMERIZE AS E | DEDUPLICATE],
   A2 [CATEGORIZE INTO L1, L2, ..., Lx |
   IMPUTE | NUMERIZE AS E | DEDUPLICATE], ...,
   An [CATEGORIZE INTO L1, L2, ..., Lx |
   IMPUTE | NUMERIZE AS E | DEDUPLICATE]
   FROM r1, r2, ..., rn
   WHERE c
\end{verbatim}

\subsection{Semantics of MQL}

The semantics we assign to each of these statements are system and implementation specific. By that we mean that each system implementing these statements will play a major role in their meaning, efficiency, accuracy and performance. For example, if they are implemented in TensorFlow, as opposed to PyTorch or R, they will demonstrate different characteristics, i.e., the predictions made by underlying TensorFlow algorithms could be different from the Pytorch or SciKit-Learn based algorithms, and the prediction accuracies may vary. We consider this aspect of MQL as somewhat similar to SQL's optimization strategies. Only difference is in the case of MQL it is more about the quality of the predictions or the semantic interpretations of data. In this paper, we do not address these issues and only focus on generic semantics we expect from each of these statements. 

For illustrative purposes, we use Kaggle's Boston housing market dataset \cite{Perera2018} as our example. This data has 506 observations with 13 continuous and 1 binary attributes stored as the file {\em bostonHomes} with the following interpretations (partial list, full list in \cite{Perera2018}):
{\small
\begin{enumerate}
\item CRIM - per capita crime rate by town
\item ZN - proportion of residential land zoned for lots over 25,000 sq.ft.
\item NOX - nitric oxides concentration (parts per 10 million)
\item DIS - weighted distances to five Boston employment centres
\item TAX - full-value property-tax rate per \$10,000
\item PTRATIO - pupil-teacher ratio by town
\item MEDV - Median value of owner-occupied homes in \$1000's
\end{enumerate}
}
Many distinct analyses for this dataset by a large number of researchers point to how a smart query processor and optimizer could exploit them to develop processing strategies to meet user needs. Our goal, however, is not to delve into processing strategies or optimization opportunities except to offer these passing comments for interested readers. Instead, we refer to the Python code segment in Fig \ref{SciMQL} that implements a linear regression model assuming a Pandas DataFrame ``{\bf df}" with columns {\em MEDV, CRIM, ZN, NOX, DIS, TAX, PTRATIO}. It splits the data into training and testing sets, standardizes the features, builds a linear regression model using SciKit-Learn, trains the model on the training set, evaluates it on the test set, and makes predictions. The number of epochs and other hyperparameters can be adjusted by a query processor for the dataset, as needed,  to meet any user specified performance threshold. Similar code segments can be generated to implement the CONSTRUCT and INSPECT statements.

\subsubsection{Query Processing}

Compared to SQL databases, ML databases and query processing are likely more nuanced, complex, and require more user involvement in library and algorithm selection, or code customization. Query processing for MQL currently needs additional instructions beyond the Python scripts similar to the one in Fig \ref{SciMQL}, and are not explained further. A file handler has been implemented to bring data to the MQL store and link with the query processor. The directory path for the Boston housing data in CSV format can be included in the Python code segment or copied into the directory where the code is running.

The MQL query for the prediction of home values using the Boston housing data in Fig \ref{mql} can be submitted in command line mode in the MQL engine for execution. In this query, the median home value is being predicted for homes in the file {\em homesNew} given a subset of features in {\em CRIM, ZN, NOX, DIS, TAX, PTRATIO}. In the plot, the {\em HomeNo} in the file {\em homesNew} is used as label.

\begin{figure}[hbt!]
   \centering
   \begin{verbatim}
    GENERATE DISPLAY OF
    PREDICTION MEDV
    OVER homesNew
    LABEL HomeNo
    FEATURES CRIM, ZN, NOX, DIS, TAX, PTRATIO
    FROM bostonHomes
\end{verbatim}
   \caption{MQL query for median home value prediction.}
   \label{mql}
\end{figure}

\subsubsection{Results}

\begin{figure}[htp!]
\centering
\subfigure[homesNew assumed table.]
{\label{homesNewzero}
\includegraphics[keepaspectratio,width=.49\textwidth]{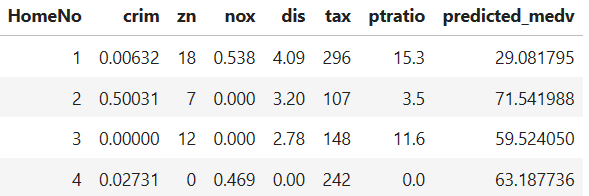}}
\subfigure[Bar plot.]
{\label{bplot}
\includegraphics[keepaspectratio,width=.49\textwidth]{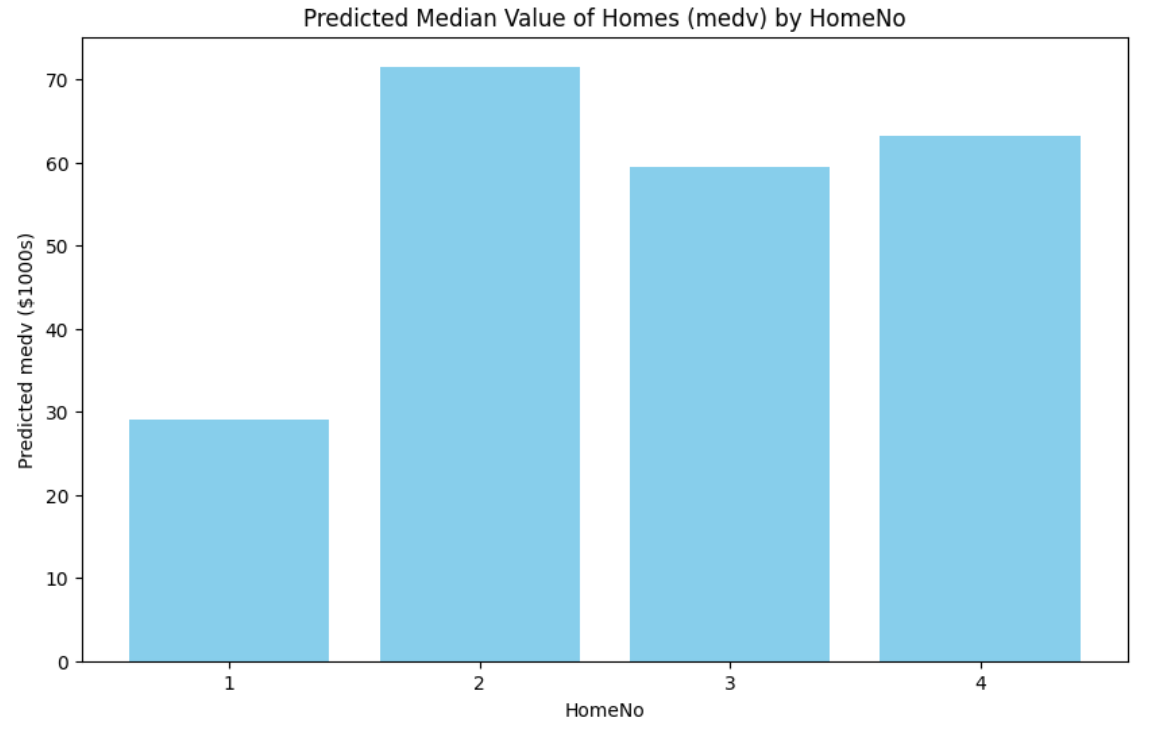}}
\subfigure[Predicted v actual plot.]
{\label{actpred}
\includegraphics[keepaspectratio,width=.49\textwidth]{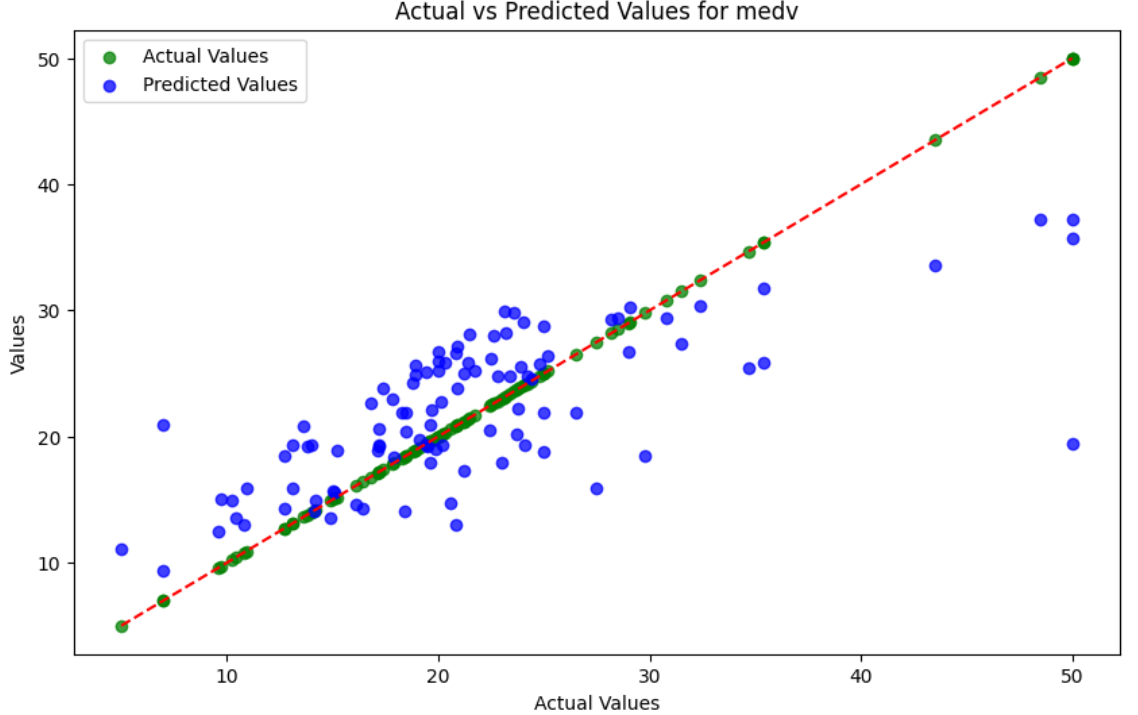}}
\caption{Predicted home median values.}
\label{plots}
\end{figure}

We assign  translational semantics to the query in Fig \ref{mql} by mapping it to the SciKit-Learn program in Fig \ref{SciMQL} for execution. In this program, we first extract the features ('CRIM', 'ZN', 'NOX', 'DIS', 'TAX', 'PTRATIO') and the target ('MEDV') from the DataFrame. We then split the data into training and testing sets using train\_test\_split. Next, we create a Linear Regression model and train it on the training data using fit. We then make predictions on the test set using predict, and evaluate the model using mean squared error (mean\_squared\_error). On execution over the slightly sparse dataset in Fig \ref{homesNew}, it produced the plot in Fig \ref{bplot} in which we assumed zero for missing values as shown in Fig \ref{homesNewzero}. If imputed values are used for missing values as shown in the code in Fig \ref{SciMQL}, predictions will be slightly different. The predicted versus actual plot is shown in Fig \ref{actpred}.

\begin{figure}[hbt!]
\centering
\begin{verbatim}
import pandas as pd
df = pd.read_csv("bostonHomes.csv")
from sklearn.model_selection import train_test_split
from sklearn.linear_model import LinearRegression
from sklearn.metrics import mean_squared_error

# Assuming you have a DataFrame 'df' with columns 'MEDV',
  'CRIM', 'ZN', 'NOX', 'DIS', 'TAX', 'PTRATIO'
# Extracting features (CRIM, ZN, NOX, DIS, TAX, PTRATIO)
  and target (MEDV)
X = df[['CRIM', 'ZN', 'NOX', 'DIS', 'TAX', 'PTRATIO']]
y = df['MEDV']

# Splitting the data into training and testing sets
X_train, X_test, y_train, y_test = train_test_split(X, y,
  test_size=0.2, random_state=42)

# Creating a linear regression model
model = LinearRegression()

# Training the model
model.fit(X_train, y_train)

# Making predictions on the test set
y_pred = model.predict(X_test)

# Evaluating the model
mse = mean_squared_error(y_test, y_pred)
print("Mean Squared Error:", mse)

# Printing the coefficients of the model
print("Coefficients:", model.coef_)

# Printing the intercept of the model
print("Intercept:", model.intercept_)

# Test set prediction
from sklearn.impute import SimpleImputer
test_samples = pd.DataFrame({
    'CRIM': [0.00632 , 0.50031, np.nan, 0.02731],
    'ZN': [18, 7, 12, 0],
    'NOX': [0.538, np.nan, np.nan, 0.469],
    'DIS': [4.09, 3.20, 2.78, np.nan],
    'TAX': [296, 107, 148, 242],
    'PTRATIO': [15.3, 3.5, 11.6, np.nan]})
imputer = SimpleImputer(strategy='median')
imputer.fit(X_train)
X_test_imputed = imputer.transform(test_samples)
X_test_imputed_df = pd.DataFrame(X_test_imputed,
  columns=['CRIM', 'ZN', 'NOX', 'DIS', 'TAX', 'PTRATIO'])
predictions = model.predict(X_test_imputed_df)
# Printing the predictions
print("Predictions for the 4 test samples:", predictions)
\end{verbatim}
\caption{SciKit-Learn Python code for MQL query in Fig \ref{mql}.}
\label{SciMQL}
\end{figure}

\begin{figure}[th!]
    \centering
    \begin{tabular}{|c|c|c|c|c|c|c|}
    \hline
       HomeNo & CRIM & ZN & NOX & DIS & TAX & PTRATIO \\
         \hline
        1 &  0.00632 & 18 & 0.538 & 4.09 & 296 & 15.3\\
        2 & 0.50031 & 7 & - & 3.20 & 107 & 3.5\\
        3 & - & 12 & - & 2.78 & 148 & 11.6\\
        4 &  0.02731 & 0 & 0.469 & - & 242 & - \\ \hline
    \end{tabular}
   \caption{Test data input table {\em homesNew}.}
   \label{homesNew}
\end{figure}

\section{Implementation Strategy}

We have implemented the MQL statements using SciKit-Learn over CSV datasets. Currently, we only support one table in CSV format in the FROM clause and no WHERE clause condition is allowed\footnote{This means that if data has to come from multiple tables, users will need to pre-process and create a single table.}. Note that these restrictions are not a limitation of the language and does not affect its expressive power.

While the current implementation is directly using Python over CSV files on Panda, a more serious implementation in PostgreSQL using User Defined Functions (UDFs) written in SQL and PL/Python \cite{RechkalovZ17a,SchuleHK020} is underway. Once completed, we should be able to compare performance of the current file based and the PostgreSQL based approaches to implementation and comment more on how these choices influence various ML query processing parameters in ways similar to P2D \cite{GrigorovGRBM23} that also takes a similar translational approach.
Opportunities also exist to decide system defaults for DISPLAY OF, data wrangling for test data (e.g., assuming zero va;ue imputation), and so on.

\subsection{Translational Semantics of MQL}

One of the most convenient and effective implementation strategies for novel languages is to map it to a fully functional and well known language. Among the popular ML frameworks such as PyTorch, Keras, TensorFlow, XGBoost, MXNet and so, SciKit Learn probably is one of the most widely used. We choose SciKit Learn for its excellent support for tabular data analysis using traditional ML tasks, and the ease of use. In this section, we develop an algorithm $\tau$ to assign a translational semantics to all MQL programs $\mathcal P$ by mapping it to a SciKit Learn program $\mathcal S$, i.e., $\tau (\mathcal P)= S$, such that $\mu(\mathcal P)\equiv \mu(S)$, where $\mu$ is a meaning function.

\begin{figure}[ht!]
   \centering
   \includegraphics[keepaspectratio,width=.49\textwidth]{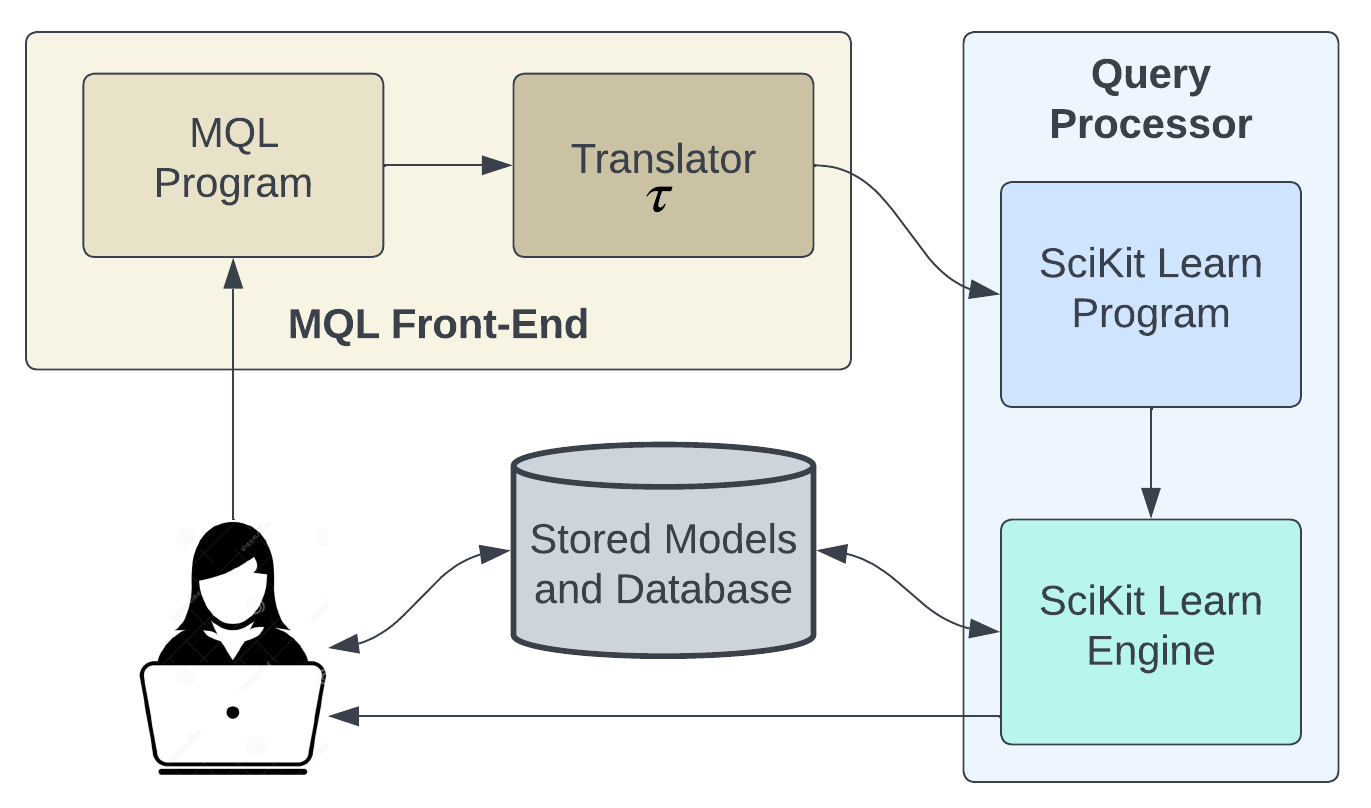}
   \caption{MQL operational model.}
   \label{model}
\end{figure}

An MQL program essentially is a sequence of one of three MQL statements - GENERATE, CONSTRUCT or INSPECT. Therefore, the meaning of a program $\mathcal P$ is the intended meaning of each of the statements in the sequence they are stated. As MQL is a context-independent language, it is easy to see that if every MQL statement $p \in \mathcal P$ could be translated into a SciKit Learn program ${\mathcal S}_p$, i.e.,  $\tau (p)={\mathcal S}_p$, such that $\mu(p)\equiv {\mathcal S}_p$, then the relationship $\mu(\mathcal P)\equiv \mu(\mathcal S)$ will hold. We therefore, design our translation function $\tau$ to map $\mathcal P$ on a case by case basis. The operational model of MQL shown in Fig \ref{model} follows this simple implementation strategy. 

It should be noted that there is an inherent dependency $\leftharpoonup$ among the MQL statements as follows. Although the semantics of the CONSTRUCT statements depends on the appropriate table representation conforming to the data types that can be adhered to by data wrangling operations using the INSPECT statements, MQL mandates that such an operation should be initiated by the user's program. In other words, invoking an INSPECT operation is not automatic even when CONSTRUCT $\leftharpoonup$ INSPECT holds, and MQL expects the program to FAIL if CONSTRUCT is not executable due to datatype errors and not corrective step using INSPECT precedes it.

On the other hand, the dependency GENERATE $\leftharpoonup$ CONSTRUCT is fully automatic. The dependency of the GENERATE statement on CONSTRUCT is manifested in two principal ways. First, when the ALGORITHM option is used and a new model generation is required as follows.
\begin{verbatim}
   GENERATE [DISPLAY OF]
   PREDICTION v [OVER s] |
   CLASSIFICATION INTO L1, L2, ..., Lp [OVER s] |
   CLUSTER OF k
   ALGORITHM AlgorithmName
   [WITH MODEL ACCURACY P]
   [LABEL B1, B2, ..., Bm]
   FEATURES A1, A2, ..., An
   FROM r1, r2, ..., rq
   WHERE c
\end{verbatim}
Or when none of the options USING or ALGORITHMS is used in the GENERATE statement indicating a default model must be generated as follows.
\begin{verbatim}
   GENERATE [DISPLAY OF]
   PREDICTION v [OVER s] |
   CLASSIFICATION INTO L1, L2, ..., Lp [OVER s] |
   CLUSTER OF k
   [WITH MODEL ACCURACY P]
   [LABEL B1, B2, ..., Bm]
   FEATURES A1, A2, ..., An
   FROM r1, r2, ..., rq
   WHERE c
\end{verbatim}
In these cases, an appropriate CONSTRUCT execution is invoked and a model is generated for use and destroyed once GENERATE ends. Note that in both cases, FEATURES option is mandatory as GENERATE needs to know which features to use.

The second way GENERATE indirectly depends on CONSTRUCT is when USING is used as follows. In this case, FEATURES need be used as a model is already built and deployed. In this case, dependency is explicitly captured using an explicit CONSTRUCT statement ahead of the GENERATE statement in the program (or executed separately and the generated model archived in the MQL system).
\begin{verbatim}
   GENERATE [DISPLAY OF]
   PREDICTION v [OVER s] |
   CLASSIFICATION INTO L1, L2, ..., Lp [OVER s] |
   CLUSTER OF k
   USING MODEL ModelName
   [WITH MODEL ACCURACY P]
   [LABEL B1, B2, ..., Bm]
\end{verbatim}

\subsection{Mapping Algorithm $\tau$}
\label{mapping}

We breakdown the mapping algorithm $\tau$ into three component algorithms driven by a main driver algorithm where the translation takes off (see Alg \ref{alg:tau}). In the Algs \ref{alg:gen} through \ref{alg:ins}, we use a function {\em generate(p)} that returns a set of descriptive properties of a syntactically correct MQL statement $p$. {\em generate} retruns the following descriptions:
\begin{itemize}
\renewcommand\labelitemi{--}
    \item StType: returns the class of statement type -- one of gen, con or ins.
    \item Model: in GENERATE, stored when USING MODEL, custom when ALGORITHM and default when none.
    \item MLtype: in GENERATE or CONSTRUCT, pred when PREDICTION, class when CLASSIFICATION and clus when CLUSTER.
    \item ModName: in GENERATE, ModelName when USING MODEL, NULL otherwise.
    \item Features: in GENERATE and CONSTRUCT, $A_1, A_2, \ldots, A_n$ when FEATURES, NULL otherwise.
    \item Display: in GENERATE, yes when DISPLAY OFF, no otherwise.
    \item Label: in GENERATE, yes when  LABEL, no otherwise.
    \item AlgName: in GENERATE and CONSTRUCT, AlgorithmName when ALGORITHM, NULL otherwise.
\end{itemize}
In the Alg \ref{alg:tau}, we call {\em generate} and assign all these descriptive features of a statement $p$ into a class $\Delta$, and pass it to the other algorithms as a decision making tool.

\SetKwComment{Comment}{/* }{ */}
\begin{algorithm}
\caption{Translator $\tau$}\label{alg:tau}
\KwData{an MQL program $\mathcal P$}
\KwResult{A SciKit Learn Script ${\mathcal S}_p$}
\Begin{
    \For{all $p\in \mathcal P$}{
        $\Delta$ $\gets$ gather($p$)\;
        \Switch{$\Delta$.StType}{
            \Case{gen}{
                call GENERATE($p$, $\Delta$)
            }
            \Case{cons}{
                call CONSTRUCT($p$, $\Delta$)
            }
            \Case{ins}{
                call INSPECT($p$, $\Delta$)
            }
        }
    \Return
    }
}
\end{algorithm}

The Alg \ref{alg:gen} for translating a GENERATE statement, uses a stored model when USING MODEL is used, otherwise it calls Alg \ref{alg:con} to use the functionalities of CONSTRUCT even though a CONSTRUCT statement is not explicitly requested. This is because when a stored model is not used in it, either a specific or custom model is requested using the ALGORITHM option, or none at all (default), which in both cases must be constructed, used and discarded. Note that, MQL has a default algorithm for each class of analysis, and it is not query or data dependent. This also means that use of a default algorithm is not always appropriate though results will be generated. To ensure analysis quality, WITH MODEL ACCURACY option can be used with default so that the system is able to find the best model for the intended analysis.

\SetKwComment{Comment}{/* }{ */}
\begin{algorithm}
\caption{GENERATE}\label{alg:gen}
\KwData{a GENERATE Statement $p$, $\Delta$}
\KwResult{A SciKit Learn Script ${\mathcal S}_p$}
\Begin{
\Switch{$\Delta$.Model}{
    \Case{stored}{
        \Switch{$\Delta$.MLType}{
            \Case{pred}{
                create ${\mathcal S}_p$ for $\Delta$.ModName for prediction of $v$ using test set $s$\;
            }
            \Case{class}{
                create ${\mathcal S}_p$ for $\Delta$.ModName for classification of $v$ using test set $s$ into classes $L_1, L_2, \ldots, L_p$\;
            }
            \Case{clus}{
                create ${\mathcal S}_p$ for $\Delta$.ModName to create $k$ clusters using test set $s$\;
            }
            }
            \If{$\Delta$.Display=yes}{
                include visualization instructions
            }
            \If{$\Delta$.Label=yes}{
                include instructions for labels $B_1, B_2, \ldots, B_m$
            }
            \Return
    }
    \Case{custom or default or best}{
        call CONSTRUCT($p$, $\Delta$)
    }
    \Return
    }
}
\end{algorithm}

\SetKwComment{Comment}{/* }{ */}
\begin{algorithm}
\caption{CONSTRUCT}\label{alg:con}
\KwData{a CONSTRUCT Statement $p$, $\Delta$}
\KwResult{A SciKit Learn Script ${\mathcal S}_p$}
\Begin{
create instructions to generate a table $T(\Delta$.Features$)$ from the FROM clause tables \Comment*[r]{includes $v$ in the feature set if $\Delta$.MLType=pred}
include instructions to divide table $T$ into $N$ training and $M$ test sets\;
\Switch{$\Delta$.Model}{
    \Case{others}{
        choose the default algorithm $A$ for $\Delta$.MLType\;
        \Switch{$\Delta$.MLType}{
            \Case{pred}{
                \eIf{$\Delta$.Model==custom}{create ${\mathcal S}_p$ for prediction of $v$ using $\Delta$.AlgName}{create ${\mathcal S}_p$ for prediction of $v$ of objects in $T$ using $A$}
            }
            \Case{class}{
                \eIf{$\Delta$.Model==custom}{create ${\mathcal S}_p$ for classification of objects in $T$ using using $\Delta$.AlgName and labels $L_1, L_2, \ldots, L_p$}{create ${\mathcal S}_p$ for classification of objects in $T$ using $A$ and labels $L_1, L_2, \ldots, L_p$}
            }
            \Case{clus}{
                \eIf{$\Delta$.Model==custom}{create ${\mathcal S}_p$ for classification of objects in $T$ using $\Delta$.AlgName for $k$ classes}{create ${\mathcal S}_p$ for classification of objects in $T$ using $A$ for $k$ classes}
            }
        }
        \Return
    }
    \Case{best}{
        \For{each algorithm $A$ of $\Delta$.MLType}{
        create ${\mathcal S}_p$ for $\Delta$.MLType of objects in $T$ using $\Delta$.AlgName\;
        }
        include instructions to choose the best model with accuracy $\geq P$
    }
    \Return
    }
}
\end{algorithm}

Alg \ref{alg:ins} implements a quicker and shortcut instruction for SQL's UPDATE statement. It supports convenient operations not directly available in UPDATE. For example, IMPUTE or DEDUPLICATE have no UPDATE counterparts.

\SetKwComment{Comment}{/* }{ */}
\begin{algorithm}
\caption{INSPECT}\label{alg:ins}
\KwData{an INSPECT Statement $p$, $\Delta$}
\KwResult{A SciKit Learn Script ${\mathcal S}_p$}
\Begin{
create instructions to generate a table $T(R)$ from the FROM clause tables \Comment*[r]{Following an SQL interpretation of the relations in the FROM clause}
    \For{every attribute $A_i$ and for every row in $T$}{
        generate instructions for categorize, impute, numerize or deduplicate
    }
\Return
}
\end{algorithm}

\section{Experimental Results}
\label{results}

\begin{figure*}
\centering
\includegraphics[scale=.7]{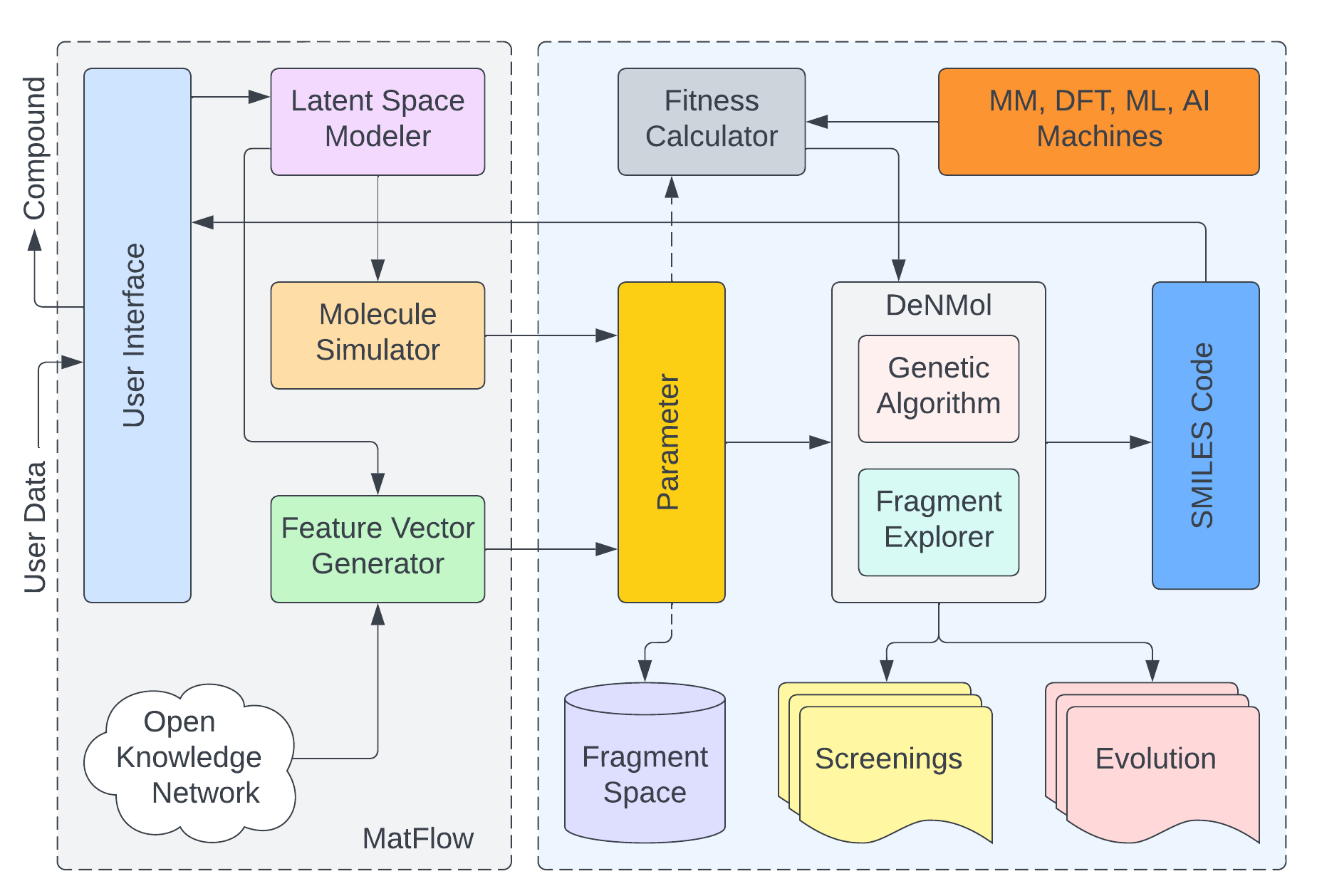}
\caption{MatFlow architecture.}
\label{matflow}
\end{figure*}

In this section, we discuss two scientific ML applications that were recently modeled using MQL in our materials design system MatFlow as shown in Fig \ref{matflow}. Our first application is a quantum dye design experiment in which we aim to discover a new quantum dye molecule with a target extinction coefficient ($\varepsilon$) higher than 250,000 $M^{-1}cm^{-1}$ using inverse ML \cite{RahmanJ2023}. The experiment involves two steps. In the first step, we use experimental data to learn the most relevant features $F$ that are significant contributors to high extinction coefficient of dyes. Then in the second step, we estimate the values of the features $F$ as a directional feature vector as candidates for our target extinction coefficient. The idea is to use these candidate features to design a novel molecule using a system such as GenUI \cite{SichoLSW21}. Our validation process involved finding Cyanine-5 (Cy5) as a possible quantum dye with an extinction coefficient $\varepsilon = 250,000 ~M^{-1}cm^{-1}$.

The second experiment is about prediction of bending modulus $\kappa$ of lipid bilayer membranes using experimental data. In this experiment, a meticulously curated dataset is used representing a large number of lipid bilayer membrane properties. We analyzed this data set using a graph convolutional neural network to generate a model for estimating the bending modulus of new experimental lipids with more than 78\% accuracy, which is much higher than existing methods.

\subsection{Quantum Dye Design}

Quantum dyes or dots are widely used in biomedical applications such as cancer detection, medical imaging, and also in studying transport mechanisms in cells, functional heterogeneity of cells, diffusion movements of membrane transport proteins, and many similar health research. Quantum dots also play a major role in solar cells, semiconductors, light emitting devices, etc. \cite{Cotta2020}. Our partner materials science lab is interested in discovering new quantum dyes that have molar extinction coefficient ($\varepsilon$) higher than or equal to 250,000 $M^{-1}cm^{-1}$ \cite{BiaggneL2022}.

However, $\varepsilon$ is not the only property of quantum dyes that are of interest. Their optical properties including high quantum yield, high brightness, high stability against photobleaching, and intermittent fluorescence signals are too in general \cite{Sun2014}, making dye discovery a multi-dimensional design problem. It is actually known that the specific characteristics of quantum dyes can vary depending on the material system (such as CdSe, PbS, or InP) and the synthesis techniques employed. Quantum dyes can be engineered with different properties by modifying their size, shape, composition, and surface functionalization, offering a wide range of possibilities for tailoring their characteristics to specific applications. Thus discovering the fact that Cy5 has $\varepsilon=250,000$ from a Google search is not sufficient \cite{Cy5}. In fact, many Alexa Fluor family of dyes have members with even higher $\varepsilon$ \cite{AlexaFluor}. 

In our experiment, we use nanoHUB \cite{Madhavan2013} quantum dye design data and set out to first determine which ML model $M$ predicts the extinction coefficient $\varepsilon$ most accurately, and learn a very small dominant set of predictive features $F$, the latent space $z$, that control $\varepsilon$. Our goal is to discover dyes that are absent in nanoHUB dataset that have $\varepsilon\geq 150,000$, such as Cyanine-5 (Cy5). We then query scientific literature, and scientific repositories such as PubChem \cite{gkac956}, ChemSpider \cite{Pence2010} or ChemDB \cite{btm341} to extract the feature values $F$ of dye molecules, and compare their predicted $\varepsilon$ by our chosen model $M$, and with experimental and literature reports, and validate the accuracy.

For our experiment, we generated a dataset, called DyeData, from nanoHUB over a 249 column feature space consisting of 8,802 dye objects. The feature space attributes represent important dye characteristics that are significant determinants of dipole moments ($\mu$) and are used to optimize coupling within dye aggregates, especially in materials research \cite{BiaggneL2022}. We have integrated and explored Deep4Chem \cite{joung2020experimental}, PhotoChem CAD 3 \cite{taniguchi2018photochemcad}, and Dyomics \cite{Dyomics}, in order to prepare our feature space. In addition, we utilized RDKit \cite{RDKit} to extract various molecular descriptors as physical properties. The context $c$ in this set up is the extinction coefficient $\varepsilon$.

The following INSPECT and CONSTRUCT statements are representative of a sequence operations we have performed in MatFlow. We created two models -- a Random Forest model and a Linear Regression Model, from which the Random Forest model was chosen for its higher model accuracy. The discovered latent feature space $F$ selected only 23 features from a total of 249 columns. A GENERATE statement is issued to predict the extinction coefficients of unknown dyes using the Random Forest Model. The results of this analysis can be found in \cite{RahmanJ2023}.

The INSPECT statement below involves two CSV files --  Chromophore.csv(Tag, MolarCoeff, $\ldots$) and High\_Extinction.csv(Tag, ShouldBe) with a join column named ``Tag". MQL supports all SQL statements and MQL statements can be used anywhere a table is expected, and vice versa. The INSPECT and SQL statements below show a sequence of data wrangling operations during the generation of the DyeData dataset. We have used about 80\% (equal to 7,040) of the total observations (8,802) in the DyeData dataset as the training data in the GENERATE statement below, and about 20\% (equal to 1,760) as the test data. 
\begin{verbatim}
   INSPECT ShouldBe NUMERIZE AS log(ShouldBe)
   FROM High_Extinction.csv;

   CREATE VIEW Temp.csv as
   SELECT Tag,
      CASE
        WHEN ShouldBe=NULL THEN ShouldBe=MolarCoeff
      END AS Epsilon
   FROM FROM Chromophore.csv LEFT OUTER JOIN
      High_Extinction.csv;

   ALTER TABLE Chromophore.csv
   DROP COLUMN MolarCoeff;

   CREATE VIEW DyeData.csv as
   SELECT *
   FROM Chromophore.csv LEFT OUTER JOIN Temp.csv;
   
   CONSTRUCT epsilonPred FOR
   PREDICTION epsilon
   USING RandomForest
   TRAIN ON 7040 TEST ON 1760
   FEATURES *
   FROM DyeData;

   GENERATE DISPLAY OF
   PREDICTION epsilon OVER TestData
   USING ALGORITHM LinearRegression
   WITH MODEL ACCURACY 80
   FEATURES *
   FROM DyeData;
   
   GENERATE DISPLAY OF
   PREDICTION epsilon OVER TestData
   USING MODEL RandonForest;
\end{verbatim}

\subsection{Membrane Bending Modulus Prediction}

Lipid bilayer membranes play an important role in the functional architecture of living cells, facilitating essential processes including the transmission of signals between and within cells and the transport of substances across the cellular barrier \cite{alberts2002genesis}. These bilayers are dynamic structures composed of molecules with hydrophilic heads and hydrophobic tails, resulting in considerable variations in both their composition and physical properties \cite{shahane2019physical}. Several factors, including molecular composition, environmental conditions, and physical state of lipids have a wide range of interactions in bilayers. These interactions make it difficult to understand and forecast the combined effects on the bilayer's properties \cite{galassi2021coupling}.

Prediction of bending modulus of lipid membranes is important in various fields such as biophysics, biochemistry, and materials science. The bending modulus is a measure of the membrane's resistance to bending or deformation and is a crucial parameter in understanding the mechanical properties and behavior of lipid membranes and has applications in lipid design, drug discovery, nanotechnology, material science and other biomedical applications. However, this application requires the use of graph neural networks which the current edition of MQL does not support. We thus implemented a PyTorch translation for the GENERATE statement so that we are able to use a GNN model created using PyTorch through the invocation of the following GENERATE statement as a one off demonstration of the versatility of MQL and that developing a multi-framework implementation of MQL is feasible.
\begin{verbatim}
   GENERATE DISPLAY OF
   PREDICTION Kappa OVER LipidTestData
   USING MODEL LipidGnn;
\end{verbatim}

\section{Discussion}
\label{discussion}

In our view, there are not too many declarative ML languages that stand at the same level as MQL. By that we mean, a language that does not require users to express analysis needs using a language more akin to procedural codes. It should be readily noticed that the languages such as Dyna, ML2SQL, sql4ml, and P6 \cite{LiM21} though possibly are more powerful and customizable, they are closer to procedural languages such as Python or C++, and thus give an appearance and flavor of imperative programming. The CREATE FUNCTION or the CREATE OPERATOR statements and the elaborate codes in Python or C++ is a significant barrier. In contrast, we hide all procedurality and offer a flavor of SQL like semantics.

As discussed in Sec \ref{related}, Dyna and P6 are both focused on optimization and visualization respectively, and thus declarativity is not their main focus. They are successful in code optimization and developing conceptual codes for easy visual analytics specification. They too are not truly comparable to MQL. We  actually agree with Gleeson \cite{Gleeson2021} and believe that declarativity should be SQL like, even for ML. Gleeson, however, encoded several ML tasks directly in PostgreSQL using ML features supported in it. For example, regression has been coded as follows where the objective is to “learn” the parameters $m$ and $c$ of a linear equation of the form $y = mx + c$ from the training data.
\begin{verbatim}
    WITH regression AS
      (SELECT regr_slope(y, x) AS gradient,
          regr_intercept(y, x) AS intercept
       FROM linear_regression
       WHERE y IS NOT NULL)
    SELECT x, (x * gradient) + intercept AS prediction
    FROM linear_regression CROSS JOIN regression
    WHERE y IS NULL;
\end{verbatim}
In this code segment, the {\em regr\_slope()} and {\em regr\_intercept()} functions are used to estimate the gradient and intercept terms, respectively. These correspond to the parameters $gradient$ and $intercept$ in the equation $y=gradient \times x + intercept$. However, this is possible only in PostgreSQL and other database engines will not recognize this code segment. In MQL, we will express the same functionality as follows, which we are able to execute on any database engine using a front-end.
\begin{verbatim}
    GENERATE DISPLAY OF
    PREDICTION y
    OVER unknown_xs
    FEATURES x, gradient, intercept
    FROM linear_regression
\end{verbatim}
In the above query, the table {\em unknown\_xs} contains the values $x$ for which $y$ needs to be predicted. The DISPLAY OF option plots a graph to show the $y$ values against each $x$ in {\em unknown\_xs}. Without the DISPLAY OF option, MQL will just compute a table with the columns $x$ and $y$. The difference obviously is, in MQL, users think in a more abstract manner and at a very conceptual level.

Our current implementation has a few drawbacks that we plan to address in MQL's future editions. The first drawback was a design choice for the first edition of MQL. In this edition, we did not include an option to generate visualization primitives for the CONSTRUCT and INSPECT statements, only GENERATE supports data visualization. But, it is necessary to allow visualization of various relationships during model building, feature selection and data wrangling. We are in the process of designing an extended set of suitable features to support data visualization. 

The second limitation of MQL is related to the quality of analysis and query processing performance. There are numerous ML frameworks and a large number of ML algorithms that are suitable for applications on a case by case basis. Therefore, it is imperative that an MQL query optimizer be developed to identify candidate algorithms most relevant to a specific ML task, data set and analytic options. In absence of such an optimizer, MQL is in risk to compromise quality of analysis or performance, or both. We hope to address this limitation soon.

\section{Conclusion}

The MQL language we have introduced has several basic strengths and advantages over other similar languages. First, its basic structure is simple and easy to understand. For example, the most basic construct for a prediction analysis is
\begin{verbatim}
    GENERATE PREDICTION f
    OVER inputData
    FEATURES f1, f2, ..., fk
    [LABEL objectName]
    FROM dataSet
\end{verbatim}
where $f$ is a feature outside the feature vector $\langle f_1, f_2, \ldots, f_k\rangle$ which must be included in the scheme of the table {\em dataSet}, and so must be {\em objectName} if LABEL option is used. The table {\em inputData} must have a scheme that includes the feature vector $\langle f_1, f_2, \ldots, f_k\rangle$ plus {\em objectName}. The statement basically requests predictions of $f$ for the objects labeled {\em objectName} with features $\{f_1, f_2, \ldots, f_k\}$. Similar comments apply to classification and clustering requests in MQL.

The algorithmic and procedural separation of MQL and its declarative semantics also offers the opportunity for selecting implementation strategies, optimization and system level customization not offered by most contemporary declarative ML languages (the few that we are aware of). Better opportunities for using large language models now emerge to map natural language queries into MQL in ways similar to SQL for a more streamlined execution, instead of mapping to archaic Python codes. While we are contemplating a PostgreSQL implementation of MQL and explore optimization strategies, its current implementation on a file based store serves as a proof of concept and demonstrates its merits. A more detailed description of MQL's implementation will be published elsewhere.

Before concluding, we would like to note that one of the goals behind designing a declarative language for ML is to explore the possibility of a conversational ML agent that is capable of understanding analysis needs and autonomously construct ML models by selecting appropriate data sets and algorithms to be able to generate responses to scientific inquiries \cite{abs-2402-07069,abs-2311-09835}. Developing an SQL like query language for ML such as MQL makes it easier to leverage decades of experience of mapping natural language to SQL \cite{ZhangDKKKS23,FanHRGCZCJZW23s}. A growing research interest is supporting exploration of ML models \cite{0001VGBS21}, and conversational systems based on ML \cite{LazzarinettiM21}. We believe that natural language to ML mapping could also help with explainable AI \cite{abs-2209-02552,abs-2309-16021}. Recent interest in using large language models (LLM) as an interface to database applications \cite{abs-2401-10506,LiHQYLLWQGHZ0LC23} opens up opportunities for LLM interfaces to scientific applications requiring ML \cite{ZhangZRLY24,ChoiMMHM23}. Our future research is aimed at developing a conversational interface for scientific inquiries in the areas of materials science and computational biology using an LLM as a front-end and a ML system at the back-end for data analysis.

\section*{Acknowledgement}

The author would like to acknowledge the help of Hasan H Rahman who carried out the experiments described in Sec \ref{results}, and also implemented the mapping algorithm $\tau$ discussed in Sec \ref{mapping}.

The experimental setup described in this article is based on a research on quantum dye design by Lan Li et al. \cite{BiaggneL2022}. The data and some of the tools were contributed by Li materials science lab which we thankfully acknowledge.  We also acknowledge the contributions of Marat Talipov's Chemistry and Biochemistry lab for providing the data and helping to design the experiment of the lipid membrane bending modulus study.

This research was partially supported by an Institutional Development Award (IDeA) from the National Institute of General
Medical Sciences of the National Institutes of Health under Grant \#P20GM103408, and a grant from the National Science Foundation \#OIA 2019609.



\end{document}